\pdfoutput=1
\documentclass[a4paper]{article}
\usepackage{INTERSPEECH2019}
\usepackage{graphicx}
\usepackage{CJKutf8}
\usepackage{subcaption}
\usepackage{url}

\title{Polyphone Disambiguation for Mandarin Chinese Using \\ Conditional Neural Network with Multi-level Embedding Features}
\name{Zexin Cai$^1$, Yaogen Yang$^{1,2}$, Chuxiong Zhang$^1$, Xiaoyi Qin$^{1,3}$, Ming Li$^1$}

\address{
  $^1$Data Science Research Center, Duke Kunshan University, Kunshan, China\\
  $^2$College of Information and Computer, Taiyuan University of Technology, Taiyuan, China  \\
  $^3$School of Electronics and Information Technology, Sun Yat-sen University, Guangzhou, China}
\email{ming.li369@dukekunshan.edu.cn}

\begin{document}

\maketitle
\begin{abstract}
This paper describes a conditional neural network architecture for Mandarin Chinese polyphone disambiguation. The system is composed of a bidirectional recurrent neural network component acting as a sentence encoder to accumulate the context correlations, followed by a prediction network that maps the polyphonic character embeddings along with the conditions to corresponding pronunciations. We obtain the word-level condition from a pre-trained word-to-vector lookup table. One goal of polyphone disambiguation is to address the homograph problem existing in the front-end processing of Mandarin Chinese text-to-speech system. Our system achieves an accuracy of 94.69\% on a publicly available polyphonic character dataset. To further validate our choices on the conditional feature, we investigate polyphone disambiguation systems with multi-level conditions respectively. The experimental results show that both the sentence-level and the word-level conditional embedding features are able to attain good performance for Mandarin Chinese polyphone disambiguation.

\end{abstract}

\section{Introduction}
The grapheme-to-phoneme (G2P) conversion is a fundamental front-end procedure in the Chinese Text-to-Speech (TTS) synthesis system, either the traditional HMM-based speech synthesis system \cite{Yao2006An, zen2007hmm} or the End-to-End speech synthesis system \cite{ping2018deep, Arik2017Deep, Arik2017Deep2, shen2018natural}. G2P typically generates a sequence of phones from a sequence of characters or graphemes \cite{rao2015grapheme}. According to the characteristics of Mandarin Chinese, there are at least 13000 commonly used Chinese characters. However, the number considerably declines to 1300 when converting the characters into phonologically allowed syllables, and even less when using Latin alphabet representation. It appears to be a suitable choice of using phonemes or syllables as units for a TTS synthesis system in a way for effective and better performance \cite{lee1989synthesis, dong2004grapheme}. While the G2P system in English TTS synthesis system aims to produce the phoneme sequences for the out-of-lexicon words, The target of a G2P system in Chinese TTS synthesis system is to convert Chinese characters to pinyins (phoneme representations with Latin alphabet in Mandarin Chinese) \cite{Yi2009Improved}. Yet one single Chinese character could have several different pronunciations in terms of different usages in a sentence. This kind of characters is called polyphonic characters. Therefore, other than the G2P system, the polyphone disambiguation system is developed to choose the correct pronunciation of a polyphonic character from several candidates based on the context. This issue is also considered to be a homograph problem, which has important applications in speech synthesis and is still not solved today \cite{rashad2010overview}.

Rule-based algorithms \cite{zhang2001disambiguation, zirong2002efficient, huang2008disambiguating} and data-driven methods \cite{Shan2017A, Liu2011Polyphone, liu2010polyphonic} are two frequently used approaches for polyphone disambiguation. The rule-based system normally chooses the pronunciation of a polyphonic character depending on the segmental text and a well-designed dictionary. However, this method requires language expertise to produce an elaborate text-analysis system for sentence segmentation as well as a robust dictionary. The dictionary today still cannot cover all the polyphonic cases.
As for data-driven approaches, mostly the polyphone disambiguation is considered as a classification task. Inspired by the approaches for G2P in English\cite{rao2015grapheme, bisani2008joint}, in recent years more research works on polyphone disambiguation using statistical machine learning techniques, like Decision Tree \cite{Liu2011Polyphone} and Maximum Entropy Model \cite{Liu2011Polyphone,Mao2007Inequality}. 

In this paper, we introduce a data-driven approach using the conditional neural network architecture \cite{isola2017image} for polyphone disambiguation. Besides using the polyphonic character embedding feature as the network input, we obtain auxiliary features from the corresponding sentence as a condition for predicting the correct pronunciation. Previous research works in polyphonic character show that: 1) The utilization of context is an effective way to solve the pronunciation disambiguation of Chinese polyphonic characters; 2) Most polyphonic word, which comprises by polyphonic character, could be used to determine the pronunciation of the polyphonic character \cite{zhang2001disambiguation}. In the light of these two characteristics, we first design an encoder module using a recurrent neural network (RNN) structure to extract the sentence-level encoding feature as the context condition. Basically, we embed each character in the sentence and adopt the bi-directional long short-term memory (BLSTM) structure to accumulate the forward context information and backward context information as the conditional feature in the sentence-level. Besides, we use a publicly released and pre-trained word-to-vector dictionary for word-level conditional vector lookup. The prediction network maps the polyphonic character embedding features and the auxiliary features to their unique pronunciation. We investigate three systems under different combinations of conditional features on a publicly available dataset. Results show that either the word-level conditional feature or the sentence-level conditional feature yields significant improvement on polyphone disambiguation.

\begin{figure*}[ht]
	\centering
	\includegraphics[scale=0.29]{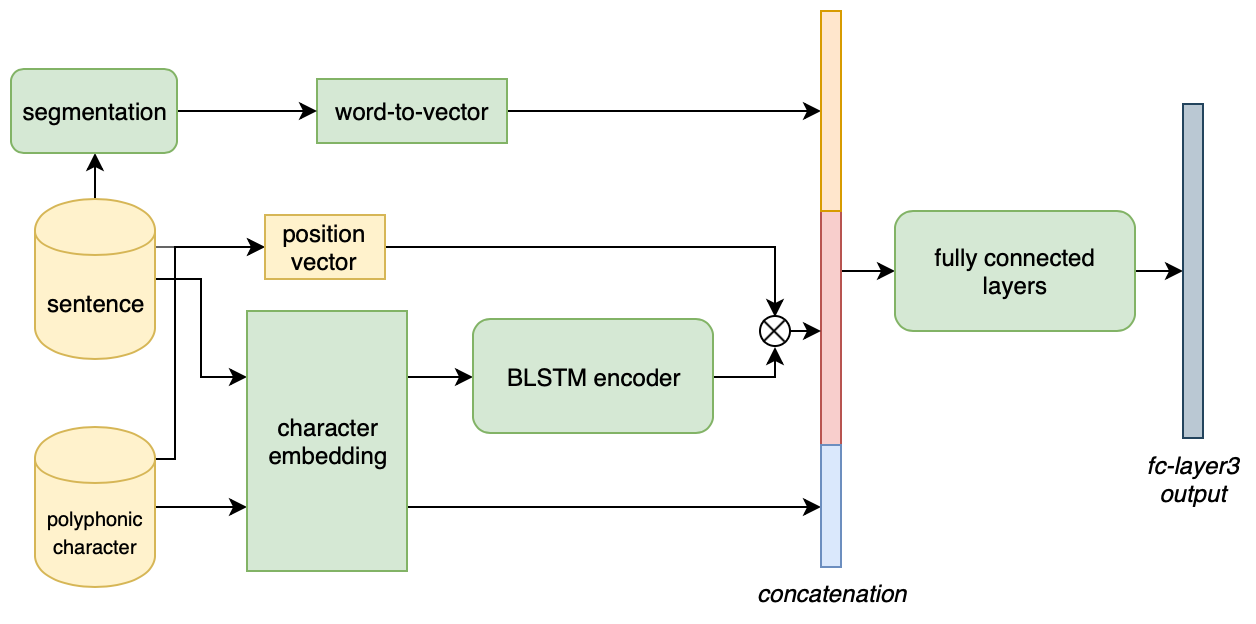}
	\caption{The network architecture of our proposed system}
	\label{fig:model}
\end{figure*}

A similar approach presented in \cite{Shan2017A} treats polyphone disambiguation as a sequence tagging task and uses BLSTM as the sequence-to-sequence generation model. However, the BLSTM structure in our proposed architecture serves as a sentence-level conditional feature extractor. Our system performs better as the accuracy reaches 94.69\% on a same evaluation set.


\begin{CJK*}{UTF8}{gbsn}
\section{Chinese Polyphonic Characters}
\label{sec:cpc}
Except for the monophonic characters in Mandarin Chinese, there are polyphonic characters that refer to those with more than one pronunciations. Specifically, we use a mapping function to formulate the conversion from a character to its corresponding pronunciations. Function $f$ is defined as follows:

\begin{equation}
	\label{equ:char2phone}
	f:C \to P
\end{equation}

where $C$ denotes the set of all Chinese characters and P denotes the set of all possible pinyins (the official romanization system for Standard Mandarin Chinese with tone information). The number of all Chinese characters is more than 80000 while the number of all pinyins is about 300. Given a character c, the output $p_c$ would be:

\begin{equation}
	\label{equ:c2phones}
	p_c = \{p_1, p_2, ..., p_n\}
\end{equation}

The character $c$ belongs to a polyphonic character when $|p_c| > 1$. For example, the Chinese character ``背" can be pronounced as either ``bei1" or ``bei4".
The number 1 to 5 denotes the tone marks since Mandarin Chinese is a tonal language. Therefore, ``bei1" and ``bei4" are two different pronunciations.

The pronunciation of a Chinese polyphonic character cannot be determined unless providing its context. 
In most cases, the pronunciation of a polyphonic character corresponds to the word that comprises the polyphonic character. The characteristics or properties of a certain word relates to the pronunciation of the polyphonic character. For example, the character ``将"  is pronounced ``jiang1" in a word where it is used as a verb or adverb, while it is pronounced ``jiang4" when the word is a noun. Specifically, ``将" is pronounced ``jiang1" in the word ``将要", which is an adverb; ``将" is pronounced ``jiang4" in the word ``大将", which is a noun. Sometimes the meaning of a word also determines the pronunciation of a polyphonic character. 
So we can rewrite formula \ref{equ:char2phone} and \ref{equ:c2phones} with:

\begin{equation}
	\label{equ:char2word}
	f:C,W_C \to P
\end{equation}

\begin{equation}
	\label{equ:c2words}
	f(c|w_c) = \{p_1, p_2, ..., p_n\}
\end{equation}
where $W_C$ is the Chinese word set containing polyphonic characters.
Normally, the pronunciation of a character $c$ given the word $w_c$ is unique. These words are called monophonic character words. 

With a well-done segmentation, we could find a unique pronunciation of the polyphonic character from word pieces except 53 special words that satisfy $|f(c|w_c)| > 1$. The pronunciation of these polyphonic character words, for example, ``朝阳" (zhao1yang2 or chao2yang2), could only be determined in a given sentence with word-level context. Moreover, when a polyphonic character stays a single character as segmental piece after a well-done segmentation, we could only utilize the context information for polyphone disambiguation. For example,
in sentence ``我不注重得与失" (Gains and losses mean nothing to me) and ``我得关注相关动态" (I have to pay attention to the relevant news), where the segmentations are ``我$\backslash$不注重$\backslash$得$\backslash$与$\backslash$失" and ``我$\backslash$得$\backslash$关注$\backslash$相关$\backslash$动态", the polyphonic word ``得" is pronounced ``de2" and ``dei3" respectively only depending on its context. 

Therefore, the pinyin of a polyphonic character is unique given its corresponding word and sentence. Which can be defined as:
\begin{equation}
	\label{equ:char2word}
	f:C,W_C,T_C \to P
\end{equation}
where $T_C$ denotes the context and $W_C$ could be $\varnothing$ if the polyphonic character word only contains a single character.

\end{CJK*}

\renewcommand\arraystretch{1.05}
\newcommand{\tabincell}[2]{\begin{tabular} {@{}#1@{}}#2\end{tabular}} 
\begin{table*}[ht]
  \caption{Detailed network structure and configurations of our proposed three systems}
  \label{model_ark}
\centering
\begin{tabular}{|cccc|c|c|c|}
\hline

\textbf{Input}	&	\multicolumn{3}{c|}{\textbf{Size}}	&	System CW	&	System CC	&	System CWC	\\ \hline
\multicolumn{1}{|c|}{character}	&	\multicolumn{3}{c|}{$B \times 1$}	&	$\surd$	&	$\surd$	&	$\surd$	\\ \hline
\multicolumn{1}{|c|}{sentence}	&	\multicolumn{3}{c|}{$B \times L$}	&	-	&	$\surd$	&	$\surd$	\\ \hline
\multicolumn{1}{|c|}{segmented word}		&	\multicolumn{3}{c|}{$B \times 1$}	&	$\surd$	&	-	&	$\surd$	\\ \hline
\textbf{Layer}	&	\textbf{Input size}	&	\textbf{Output size}		&	\textbf{Configurations}	&	&	&	\\ \hline
\multicolumn{1}{|c|}{char-embedding}		&	\multicolumn{1}{c|}{$B \times 1$}	&	\multicolumn{1}{c|}{$B \times 100$}	&	-	&	$\surd$	&	$\surd$	&	$\surd$	\\ \hline
\multicolumn{1}{|c|}{text-embedding}		&	\multicolumn{1}{c|}{$B \times L$}	&	\multicolumn{1}{c|}{$B \times L \times 100$}	&	-	&	-	&	$\surd$	&	$\surd$	\\ \hline
\multicolumn{1}{|c|}{word2vec}	&	\multicolumn{1}{c|}{$B \times 1$}	&	\multicolumn{1}{c|}{$B \times 200$}	&	- 	& 	$\surd$	&	-	&	$\surd$	\\ \hline
\multicolumn{1}{|c|}{BLSTM encoder}	&	\multicolumn{1}{c|}{$B \times L \times 100$}	&	\multicolumn{1}{c|}{$B \times L \times 512$}	&	\tabincell{c}{dropout rate: 0.1\\fw-lstm size: 256\\bw-lstm size: 256}	&	-	&	$\surd$	&	$\surd$	\\ \hline
\multicolumn{1}{|c|}{concatenation}	&	\multicolumn{1}{c|}{-}	&	\multicolumn{1}{c|}{$B \times C$}	&	-	&	$C=300$	&	$C=612$	&	$C=812$	\\ \hline
\multicolumn{1}{|c|}{fc-layer1}	&	\multicolumn{1}{c|}{$B \times C$}	&	\multicolumn{1}{c|}{$B \times 512$}	&	\tabincell{c}{activation: RELU\\dropout rate: 0.1}	&	$C=300$	&	$C=612$	&	$C=812$	\\ \hline
\multicolumn{1}{|c|}{fc-layer2}	&	\multicolumn{1}{c|}{$B \times 512$}	&	\multicolumn{1}{c|}{$B \times 1024$}	&	\tabincell{c}{activation: RELU\\dropout rate: 0.1}	&	$\surd$	&	$\surd$	&	$\surd$	\\ \hline
\multicolumn{1}{|c|}{fc-layer3}	&	\multicolumn{1}{c|}{$B \times 1024$}		&	\multicolumn{1}{c|}{$B \times 285$}	&	\tabincell{c}{activation: None\\dropout rate: 0}	&	$\surd$	&	$\surd$	&	$\surd$	\\ \hline
\end{tabular}
\end{table*}

\section{Method}
\label{sec:methods}
Different from the traditional grapheme-to-phoneme (G2P) conversion, the polyphone disambiguation is considered as a classification problem. Specifically, the polyphone disambiguation system converts a polyphonic character to its corresponding pinyin. Our proposed system is shown in Figure \ref{fig:model}. In terms of the characteristics and properties of the polyphonic character outlined in the previous section, we explore the word-level conditional feature and sentence-level conditional feature from the input sentence to help predict the pinyin. 

\subsection{Embedding}
It is a typical case to convert the characters and sentence into vectors when applying the neural network approaches. First, we initialize an embedding table with size $N_c \times D_c$ for character-to-vector lookup, where $N_c$ is the number of all characters and $D_c$ denotes the character embedding size. Before sentence embedding, we pad the character sequence to the max length of the sentences in each batch with the symbol "$|$"\ , which does not originally appear in the Chinese text to avoid any possible conflicts, without affecting the performance of the encoder module. Hence, the sentence embedding size would be $B \times L_{max} \times D_c$. $B$ refers to the batch size in the network training phase and evaluation phase.

As for word-level conditional feature, we segment the input sentence into word pieces and obtain a word sequence from a pre-trained word-to-vector lookup table. We only choose the word that comprises the polyphonic character from the segmental sentence for table lookup. 

\subsection{Bidirectional LSTM Encoder}
The Recurrent Neural Network (RNN) architecture has an elegant way of dealing with sequential problems since it is able to embody correlations between samples in the sequence\cite{mikolov2010recurrent, mikolov2011extensions, graves2013speech} . In order to address the exploding and vanishing gradient problems in RNN, the long short-term memory (LSTM) structure was proposed and successfully kept track of arbitrary long-term dependencies between the elements in the input sequences \cite{hochreiter1997long}. The LSTM structure is widely used in addressing sequence-to-sequence problems since the fundamental module of Encoder-Decoder architecture has been proven to be very effective. Encoder and decoder are the two main modules in this kind of architecture. The encoder aims to encode the input sequence to a fixed-length vector for the decoder to map the vector back into an output sequence. The Encoder-Decoder architecture demonstrated state-of-the-art performance in researches on the sequential problem including text translation, speech recognition, speech synthesis \cite{shen2018natural,cho2014properties,chan2016listen}. 

In our case, we adopt the encoder component to extract the sentence-level conditional feature from the character sequence. One problem of the LSTM is that it is unidirectional. The LSTM can only accumulate the information of the sequence in one direction. This paper uses the bi-directional LSTM \cite{schuster1997bidirectional} to accumulate the character correlations in both directions. Different from \cite{Shan2017A}, which uses BLSTM as a sequence-to-sequence generator to obtain the pinyin of a polyphonic character, we use BLSTM as an encoder to extract the sentence-level conditional feature. Suppose the character embedding sequence is $T_c=[c_1^{\top}, c_2^{\top}, c_3^{\top}, \cdots c_t^{\top}, \cdots, c_L^{\top}]$, we obtain the sentence-level conditional feature $C_{sentence}^c$ as follows:

\begin{eqnarray}
	\label{posvec}
	C_{sentence}^c &=& z_c^t \cdot BLSTM(T_c) \\
	&=& z_c^t \cdot concat(fw,bw)
\end{eqnarray}

where $z_c^t$ is the one-hot vector that denotes the polyphonic character position in the character sequence, $fw$ is the output sequence of the forward LSTM given input $T_c$ and $bw$ is the output sequence of backward LSTM. Operation $concat()$ means concatenation. The size of $z_c^t$ is $1 \times L$ and the size of $BLSTM(T_c)$ is $L \times (D_{fw} + D_{bw})$, where $D_{fw}$ and $D_{bw}$ are the output size of forward LSTM and backward LSTM.

\subsection{Prediction Network}
After concatenating the word-level conditional feature, sentence-level conditional feature and the polyphonic character embedding vector, we use several fully-connected layers following a $softmax$ layer for classification to predict the corresponding pinyin. The length of output vector equals the number of all possible pinyins in our polyphonic character database.

\renewcommand\arraystretch{1.05}
\begin{CJK*}{UTF8}{gbsn}
\begin{table*}[t]
  \caption{Experimental performance of six individual systems}
  \label{result}
  \centering
  \begin{tabular}{ | c | c | c | c | c | c | c | c | c | c | }
  \hline
	\tabincell{c}{polyphonic \\ character} & \tabincell{c}{high-fre \\ pinyin} & \tabincell{c}{low-fre \\ pinyin} & \tabincell{c}{high-fre \\ pinyin rate} & 
	\tabincell{c}{\cite{Shan2017A} \\0 word \\ accuracy} & \tabincell{c}{\cite{Shan2017A} \\1 word \\accuracy} & \tabincell{c}{\cite{Shan2017A} \\2 words \\ accuracy} & \tabincell{c}{ Our \\ System CW \\ accuracy} & 
	\tabincell{c}{Our \\ System CC \\ accuracy} & \tabincell{c}{Our \\ System CWC \\accuracy }\\ \hline
	传 & chuan2 & zhuan4 & 86.11\% & 88.89\% & 88.89\% & 88.89\% & \textbf{94.44\%} & 91.67\% & \textbf{94.44\%} \\ \hline
	只 & zhi3 & zhi1 & 70.59\% & 93.38\% & 93.38\% & \textbf{94.12\%} & 86.03\% & \textbf{94.12\%} & 93.38\% \\ \hline
	处 & chu4 & chu3 & 81.58\% & 91.67\% & 86.11\% & 91.67\% & \textbf{94.44\%} & 88.89\% & \textbf{94.44\%} \\ \hline
	少 & shao3 & shao4 & 93.98\% & 96.24\% & 95.49\% & 96.24\% & 96.24\% & 93.23\% & \textbf{96.99\%} \\ \hline
	为 & wei2 & wei4 & 59.28\% & 59.58\% & 82.38\% & 82.9\% & 62.69\% & 81.35\% & \textbf{86.53\%} \\ \hline
	藏 & zang4 & cang2 & 54.55\% & 80\% & 79.09\% & \textbf{85.45\%} & 80\% & 76.36\% & 81.82\% \\ \hline
	overall & - & - & 86.77\% & 89.13\% & 91.96\% & 91.6\% & 92.44\% & 92.44\% & \textbf{94.69\%} \\ \hline
  \end{tabular}

\end{table*}
\end{CJK*}

\section{Experimental Results}
\label{sec:results}

\subsection{Polyphonic Character Database}
\label{sec:database}
For training and evaluating our proposed polyphone disambiguation systems, we use a publicly available dataset from Beijing Data-Baker Science and Technology Ltd which contains 150 frequently used polyphonic characters and their 151585 corresponding sentences. We divide the corpus into a training set with 140794 sentences and an evaluation set with 10791 sentences \cite{biaobei}. The evaluation set is 7\% of most character-pinyin pairs and well split in with all samples concerning every different character-pinyin pair. However, some character-pinyin pairs are less than 15 samples and 20\% of those pairs are split for evaluation set. 
\subsection{Word-level Embedding}
The text segmentation tool we used for obtaining the word and phrase pieces is Jieba\footnote{https://github.com/fxsjy/jieba} Python package.
We use the released Tencent AI Lab Embedding Corpus for Chinese Words and Phrases as the pre-trained word vector lookup table. This corpus provides 200-dimensional vector representations for over 8 million Chinese words and phrases embedding. This pre-trained model was trained with large-scale text collected from news, webpages, and novels using directional skip-gram \cite{song2018directional, tecentEmbbeding}.

\subsection{System Setup}
\label{system_setup}
Our experimental setup is shown in Table \ref{model_ark}. In this paper,  we propose three different systems according to different combinations of sentence-level conditional feature and word-level conditional feature:
\begin{itemize}
\setlength{\itemsep}{1pt}
\setlength{\parsep}{0pt}
\setlength{\parskip}{0pt}
	\item System WC: concatenate the word-level conditional feature and polyphonic character embedding vector for predicting the corresponding pinyin.
	\item System CC: concatenate the sentence-level conditional feature and polyphonic character embedding vector for predicting the corresponding pinyin.
	\item System CWC: adopt both the word-level and sentence-level feature as the condition for polyphonic character embedding vector to predict pinyin. 
\end{itemize}
As shown in Table \ref{model_ark}, we adopt a single BLSTM encoder with input size $ B \times L \times 100 $, where B denotes the batch size and L denotes the length of the padding character sequences in a batch. Both the forward LSTM size and backward LSTM size is 256. The dropout rate of LSTM is set to 0.1 to avoid overfitting \cite{srivastava2014dropout}. For the prediction module, we adopt three fully connected layers with size 512, 1024 and 285 respectively. The output size 285 is equal to the number of all possible pinyins in this polyphonic character database. The activation function of the first two fully connected layers is RELU. We use dropout layers with 0.1 dropout rate after the first two fully connected layers in the training phase.  Depending on different input conditional features, the input size $C$ of prediction network can be 300, 612 and 812, respectively. The concatenated feature comprises the 100-dimensional polyphonic character embedding feature and the conditional feature (200-dimensional word-level embedding vector, 512-dimensional sentence-level encoding feature). We adopt the stochastic gradient descent (SGD) algorithm with an initial learning rate 0.1 for the training phase. The learning rate decays every 600 epochs exponentially to $10^{-4}$.

We also implement three baseline systems following \cite{Shan2017A} for comparison. We strictly follow the approach describes in \cite{Shan2017A} which adopts two LSTM layers with size 512 and the NLPIR toolkit \cite{zhou2003nlpir} for POS tagging on the text. We use the polyphonic character database described in section \ref{sec:database} for training and evaluating since we do not have the personal labelled data used in \cite{Shan2017A}. The approach presented in \cite{Shan2017A} had compared with other polyphone disambiguation approaches and shown that it reaches a better performance. Three systems regarding the different length of context inputs are implemented for comparison in our experiment:

\begin{itemize}
\setlength{\itemsep}{1pt}
\setlength{\parsep}{0pt}
\setlength{\parskip}{0pt}
	\item 0 word, not using the context information
	\item 1 word, using the past and future context information in 1 word, 3 words in total
	\item 2 words, using the past and future context information in 2 words, 5 words in total
\end{itemize}

\begin{CJK*}{UTF8}{gbsn}
\subsection{System evaluation results}
Table \ref{result} gives the performance regarding six systems illustrated in Section \ref{system_setup}. We list several polyphonic characters for comparison. 
Both the word-level conditional feature and sentence-level conditional feature help improve the disambiguation system as the performance of System CW and System CC reach 92.44\%. By concatenating the word embedding conditional feature and sentence-level conditional feature as auxiliary condition, our approach achieves the best performance with 94.69\% accuracy, which is 2.73\% higher than the best system in \cite{Shan2017A}.  

\end{CJK*}

\section{Conclusions}

In this paper, we propose a data-driven approach using conditional neural network architecture for Mandarin Chinese polyphone disambiguation. We explore sentence-level encoding vector as a condition as well as the word-level vector obtained from a pre-trained word-to-vector lookup table. Results show that the sentence-level conditional feature obtained from a single bi-directional long short-term memory structure is very useful for polyphone disambiguation. Both the word-level conditional feature and sentence-level conditional feature help improve the disambiguation system as the accuracy reaches 92.44\%. Finally, The final system achieves significant improvements with 94.69\% accuracy on the evaluation set.

\section{Acknowledgments}
This research was funded in part by the National Natural Science Foundation of China (61773413), Natural Science Foundation of Guangzhou City
(201707010363), Six Talent Peaks project in Jiangsu Province (JY-074), Science and Technology Program of Guangzhou City (201903010040). 

\vfill\pagebreak

\bibliographystyle{IEEEtran}

\bibliography{cai.bib}

\end{document}